\pdfoutput=1

\documentclass[11pt]{article}
\usepackage[final]{acl}
\usepackage{times}
\usepackage{latexsym}
\usepackage[T1]{fontenc}
\usepackage[utf8]{inputenc}
\usepackage{microtype}
\usepackage{inconsolata}

\usepackage{xspace}
\usepackage{multirow}
\usepackage{booktabs}
\usepackage{bm}
\usepackage{amsmath}
\usepackage{amsfonts}
\usepackage{hyperref}
\usepackage{graphicx}
\usepackage{framed}
\usepackage{enumitem}
\usepackage{pifont}
\usepackage{CJKutf8}
\usepackage{marvosym}
\usepackage{lipsum}
\usepackage{pmboxdraw}
\usepackage{caption}
\usepackage{subcaption}
\usepackage{epigraph}
\usepackage{longtable}

\newcommand{\eg}{\emph{e.g.,}\xspace}

\newcommand{\etc}{\emph{etc}}
\newcommand{\ignore}[1]{}

\makeatletter
\def\@fnsymbol#1{}
\makeatother

\title{LLMBox: A Comprehensive Library for Large Language Models}

\author{
Tianyi Tang\textsuperscript{\rm{1*}}\thanks{*\ Co-leading the project.}, 
Yiwen Hu\textsuperscript{\rm{1*}}, \\
\textbf{Bingqian Li}\textsuperscript{\rm{1$\dagger$}}\thanks{$\dagger$\ Equal Contribution. Ordered by name.}, 
\textbf{Wenyang Luo}\textsuperscript{\rm{1$\dagger$}}, 
\textbf{Zijing Qin}\textsuperscript{\rm{3$\dagger$}}, 
\textbf{Haoxiang Sun}\textsuperscript{\rm{2$\dagger$}}, 
\textbf{Jiapeng Wang}\textsuperscript{\rm{1$\dagger$}}, \\
\textbf{Shiyi Xu}\textsuperscript{\rm{1}}, 
\textbf{Xiaoxue Cheng}\textsuperscript{\rm{1}}, 
\textbf{Geyang Guo}\textsuperscript{\rm{1}}, 
\textbf{Han Peng}\textsuperscript{\rm{1}}, 
\textbf{Bowen Zheng}\textsuperscript{\rm{1}}, \\
\textbf{Yiru Tang}\textsuperscript{\rm{1}}, 
\textbf{Yingqian Min}\textsuperscript{\rm{1}}, 
\textbf{Yushuo Chen}\textsuperscript{\rm{1}}, 
\textbf{Jie Chen}\textsuperscript{\rm{1}}, 
\textbf{Yuanqian Zhao}\textsuperscript{\rm{1}}, \\
\textbf{Luran Ding}\textsuperscript{\rm{1}}, 
\textbf{Yuhao Wang}\textsuperscript{\rm{1}}, 
\textbf{Zican Dong}\textsuperscript{\rm{1}}, 
\textbf{Chunxuan Xia}\textsuperscript{\rm{1}}, \\
\textbf{Junyi Li}\textsuperscript{\rm{1}}, 
\textbf{Kun Zhou}\textsuperscript{\rm{2}}, 
\textbf{Wayne Xin Zhao}\textsuperscript{\rm{1 \Letter}}\thanks{\textsuperscript{\Letter}\ Corresponding author.}, 
\textbf{Ji-Rong Wen}\textsuperscript{\rm{1,2}} \\
\textsuperscript{1} Gaoling School of Artificial Intelligence, Renmin University of China \\
\textsuperscript{2} School of Information, Renmin University of China \\
\textsuperscript{3} School of Computer Science and Technology, Xidian University \\
\texttt{steventianyitang@outlook.com \ \ huyiwenwen@foxmail.com \ \ batmanfly@gmail.com} \\
}

\begin{document}
\maketitle
\begin{abstract}

To facilitate the research on large language models (LLMs), this paper presents a comprehensive and unified library, \textbf{LLMBox}, to ease the development, use, and evaluation of LLMs. This library is featured with three main merits: (1) \emph{a unified data interface} that supports the flexible implementation of various training strategies, (2) \emph{a comprehensive evaluation} that covers extensive tasks, datasets, and models, and (3) \emph{more practical consideration}, especially on user-friendliness and efficiency.  With our library,  users can easily reproduce existing methods, train new models, and conduct comprehensive performance comparisons. To rigorously test LLMBox, we conduct extensive experiments in a diverse coverage of evaluation settings, and experimental results demonstrate the effectiveness and efficiency of our library in supporting various implementations related to LLMs. The detailed introduction and usage guidance can be found at \url{https://github.com/RUCAIBox/LLMBox}.

\end{abstract}

\section{Introduction} \label{sec:intro}

Recent years have witnessed the rapid progress of large language models~(LLMs)~\cite{zhao-etal-2023-survey}. In the research community, great efforts have been devoted to the release of well-trained LLMs, the design of special tuning and inference methods, and the conduct of systematic capacity evaluation.
However, the reproducibility and fair comparison of existing studies should still be emphasized, since they are mostly developed in different ways or frameworks. Without the standardized and unified implementation, it would take substantial efforts to reproduce or improve upon existing research work.  

Considering the above issue, in this paper, we present a comprehensive library, called \textbf{LLMBox}, for easing the development, use, and evaluation of LLMs. In particular, our library focuses on building a comprehensive and unified framework (including training, inference, and evaluation) for better supporting LLM-based research and applications. Although there are already several open-source libraries for LLMs~\cite{kwon-etal-2023-efficient,eval-harness,llama-factory}, they typically focus on a certain or some stage(s) of LLMs (either pre-training or fine-tuning) or conduct the training pipeline of LLMs in a separate way. Moreover, they can seldom support comprehensive and unified evaluation of various LLMs. 

In order to better facilitate research on LLMs, LLMBox introduces a series of new features for the library design, which can be summarized into three major aspects below:

$\bullet$ \emph{Unified data interface.}
We design a unified data interface to encapsulate different formats of training data, including both plain texts and instruction data. With this interface, LLMBox can flexibly support the implementation of various strategies, such as dynamic mixture proportion~\cite{Xie-etal-2023-doremi} and combined training with pre-training and instruction data~\cite{zeng-etal-2022-glm}. 
Furthermore, we extensively support mainstream training methodologies, including parameter-efficient tuning (\eg LoRA~\cite{hu-etal-2021-lora}) and alignment tuning (\eg PPO~\cite{schulman-etal-2017-proximal}).

$\bullet$ \emph{Comprehensive evaluation.}
To support a comprehensive comparison of LLMs' performance, our library encompasses {18} downstream tasks alongside {56} datasets. Beyond the common benchmarks such as MMLU~\cite{hendrycks-etal-2021-measuring} and GSM8K~\cite{cobbe-etal-2021-training}, our framework also extends the support for probing LLMs' {advanced capabilities}: human alignment, hallucination detection, instruction following, \etc. Furthermore, LLMBox integrates a variety of publicly available LLMs and commercial APIs, offering a convenient platform for holistic evaluation.

$\bullet$ \emph{More practical considerations.} 
To be user-friendly, LLMBox is designed to provide an easy-to-use pipeline, enabling users to quickly start with minimal commands. We introduce a \emph{GPU calculator} 
to help users determine the minimum GPU resources necessary for training. 
To be efficient, we propose a novel \emph{prefix caching} strategy for inference and a \emph{packing} strategy for training. 
Remarkably, given the LLaMA (7B) model,  our library can perform inference on 
the entire MMLU benchmark within six minutes on a single A800 GPU and completes instruction tuning with 52K instances on eight A800 GPUs in ten minutes.

An additional feature is that  LLMBox is closely aligned with our prior survey paper on LLMs~\cite{zhao-etal-2023-survey}. 
This is particularly useful for beginners, enabling the learning of basic knowledge and practice of LLMs through integrating the survey paper and the associated library.

In what follows, we will first introduce the training framework of our library in Section~\ref{sec:design-training}, then detail the utilization and evaluation parts in Section~\ref{sec:design-utilization}, and showcase how to use our library in Section~\ref{sec:usage}. After that, we will conduct the experiments to verify the reliability of our LLMBox in Section~\ref{sec:exp}, and  conclude the  paper in Section~\ref{sec:conclusion}.

\begin{figure*}
    \centering
    \includegraphics[width=\linewidth]{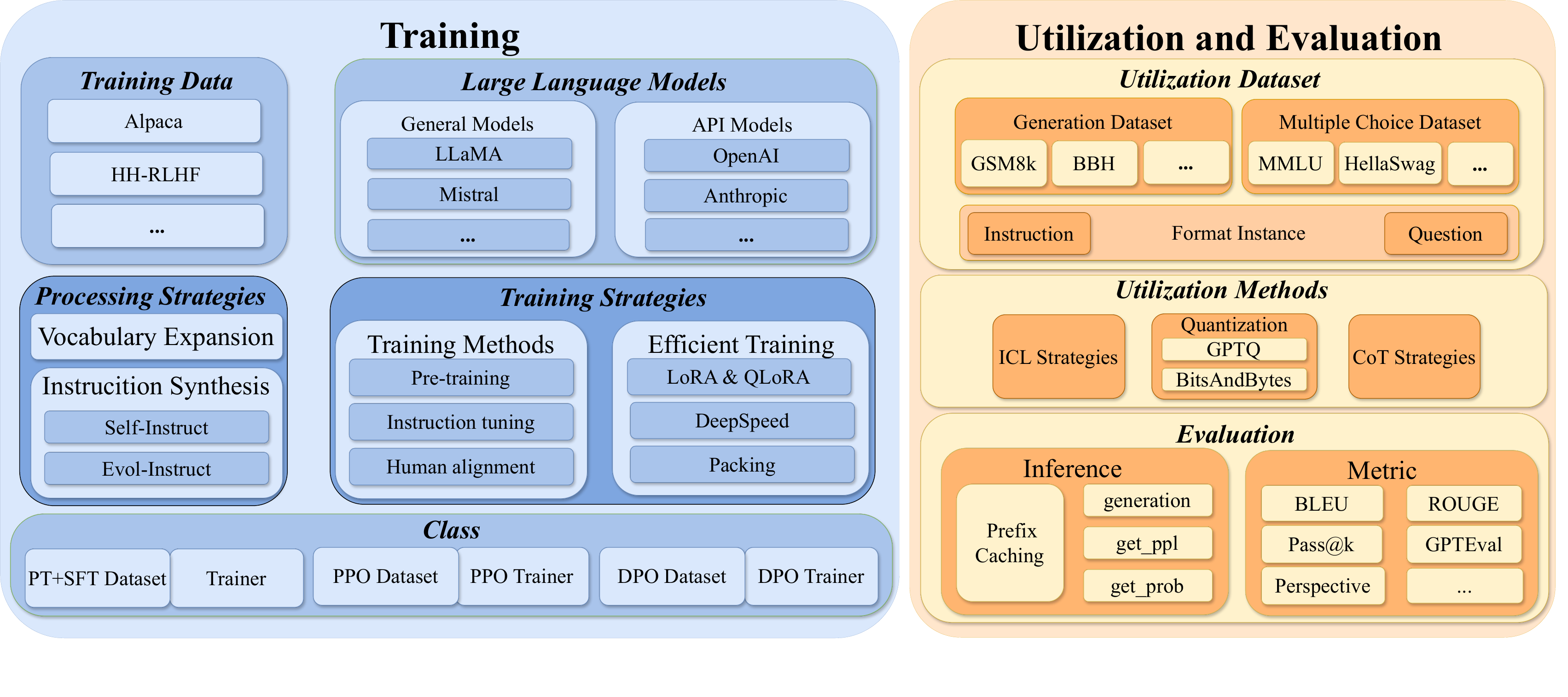}
    \caption{The overall framework of our LLMBox, supporting the training, utilization and evaluation of LLMs.}
    \label{fig:framework}
\end{figure*}

\section{Training} \label{sec:design-training}
The training process is a crucial step for the development of LLMs.
However, it typically needs massive detailed designs considering both efficiency and effectiveness, and also often faces intractable problems when adapting into new domains or meeting special needs.
To facilitate easy training of LLMs, we integrate various training methods and resources in our library, to unify and simplify their usage.
Besides, we provide suggestions for GPU usage tailored to different training requirements.

\subsection{LLM Training}
In our LLMBox, we develop a unified architecture to encapsulate important training methods in developing LLMs, and implement efficient training strategies to support training on limited computing resource.
The overall framework of LLMBox is illustrated in Figure~\ref{fig:framework}.

\paragraph{Key Training Methods.}
In our LLMBox, we integrate massive functionalities to support the following four training processes:

$\bullet$ \emph{Pre-training.}
Our LLMBox supports pre-training LLMs from scratch or continually pre-training using corpora in specific languages or specialized domains. 
For continually pre-training, LLMBox supports expanding the vocabulary to facilitate the adaptation of LLMs to new fields. 

$\bullet$ \emph{Instruction tuning.}
LLMBox integrates ten commonly-used datasets for supporting instruction-tuning, covering NLP task (\eg FLAN v2~\cite{chung-etal-2022-scaling}), daily chat (\eg ShareGPT~\cite{eccleston-etal-2023-sharegpt}), and synthetic datasets (\eg Alpaca-52K~\cite{taori-etal-2023-alpaca}).
Additionally, we integrate three methods to synthesize or rewrite instructions, namely Self-Instruct~\cite{wang-etal-2023-consistency}, Evol-Instruct~\cite{xu-etal-2023-wizardlm}, and topic diversifying~\cite{YuLan-Chat}. 
Based on the above datasets, we specially design unified dataset class, which can automatically preprocess these datasets into a unified format for training LLMs, and provide flexible interfaces for users to adjust the settings about the data (\eg data mixture proportion).

$\bullet$ \emph{Human alignment.}
To enhance the alignment of LLMs with human values, we incorporate both the widely-used RLHF method PPO~\cite{schulman-etal-2017-proximal} and non-RL approach DPO~\cite{rafailov-etal-2023-direct}. Besides, LLMBox also integrates several preference datasets, including HH-RLHF~\cite{bai-etal-2022-training} and SHP~\cite{ethayarajh-etal-2022-understanding}.

\paragraph{Efficient Training Strategies.}
We also integrate several widely-used efficient training strategies or libraries, to support training LLMs with limited computing resources.

$\bullet$ \emph{LoRA and QLoRA.}
LLMBox integrates the lightweight module LoRA~\cite{hu-etal-2021-lora} to facilitate the different training methods of LLMs in resource-constrained environments. We also encapsulate QLoRA~\cite{dettmers-etal-2023-qlora} in LLMBox, which performs quantization on LoRA for further reducing its used GPU memory.

$\bullet$ \emph{DeepSpeed.}
Our LLMBox library is based on the distributed training library DeepSpeed~\cite{rasley-etal-2020-deepspeed}, which includes a range of training optimization strategies for efficient training LLMs, including zero redundancy optimizer~(ZeRO)~\cite{rajbhandari-etal-2020-zero}, gradient checkpointing~\cite{chen-etal-2016-training}, FlashAttention~\cite{dao-etal-2022-flashattention}, \etc.

$\bullet$ \emph{Packing.}
We implement the packing strategy~\cite{raffel-etal-2020-exploring,touvron-etal-2023-llama2} to enhance training efficiency. During pre-training, we concatenate all tokens into a long sentence and then split it to multiple sentences with the max length. For instruction-tuning, we concatenate all instructions as a long multi-turn conversation, and then break it into multiple conversations approaching to the maximum length constraint.
Through minimizing paddings, we can optimize memory efficiency while maintaining model performance.

\subsection{Training Suggestions}
In practice, it is necessary for users to estimate the hardware requirements for training LLMs.
Based on our LLMBox, we meticulously analyze GPU memory consumption throughout the model training process, by fully considering the impacts of parameters, gradients, optimizer states, and activation value~\cite{rajbhandari-etal-2020-zero,ren-etal-2021-zero,korthikanti-etal-2023-reducing}. 
We further introduce a ``GPU memory calculator'' to aid users in determining the minimal GPU requirements across LLMs with different parameter scales.

By merging the above strategies to reach efficiency\footnote{For the training settings, we utilize data parallelism, ZeRO-3, gradient checkpointing, and FlashAttention.}, the memory consumption of each GPU can be roughly estimated by the equation:
\begin{equation}
    \frac{16p}{n} + (12+2l)bsh + 12bsv, \label{eq-gpu}
\end{equation}
where $p$ represents the total number of parameters, and $n$, $l$, $b$, $s$, $h$, $v$ stand for the number of GPUs, number of layers, batch size, sequence length, hidden size, and vocabulary size, respectively.
Taking the training of LLaMA-2~(7B) ($l=32, s=4096, h=4096, v=32000$) as an example, we employ two A100~(80G) GPUs ($n=2$) with a batch size of $b=8$. 
By using Eq.~\ref{eq-gpu} with the above configuration, we can estimate an approximate GPU memory usage of 71.42GB per unit.
As shown in Table~\ref{tab:GPU}, we extrapolate the minimum GPU requirements using Eq.~\ref{eq-gpu} for different model sizes across varying training settings, to help users for selecting proper GPU resources.
For other special training settings, we invite users to utilize the calculator available on our library\footnote{\url{https://github.com/RUCAIBox/LLMBox/blob/main/training/gpu_calculator.py}}.

\begin{table}
\centering
\small
\begin{tabular}{c|ccccc}
\toprule
 & \textbf{DDP} & \textbf{ZeRO-3} & \textbf{LoRA} & \textbf{QLoRA} \\
\midrule
\multirow{2}{*}{\textbf{1.3B}} & 1 A100 & 1 A100 & 1 A100 & 1 A100 \\
& 1 A6000 & 1 A6000 & 1 A6000 & 1 A6000 \\
\midrule[0.3pt]
\multirow{2}{*}{\textbf{2.7B}} & 1 A100 & 1 A100 & 1 A100 & 1 A100 \\
& N/A & 2 A6000 & 1 A6000 & 1 A6000 \\
\midrule[0.3pt]
\multirow{2}{*}{\textbf{6.7B}} & N/A & 2 A100 & 1 A100 & 1 A100 \\
& N/A & 3 A6000 & 1 A6000 & 1 A6000 \\
\midrule[0.3pt]
\multirow{2}{*}{\textbf{13B}} & N/A & 3 A100 & 1 A100 & 1 A100 \\
& N/A & 5 A6000 & 1 A6000 & 1 A6000 \\
\midrule[0.3pt]
\multirow{2}{*}{\textbf{30B}} & N/A & 8 A100 & 1 A100 & 1 A100 \\
& N/A & 12 A6000 & 2 A6000 & 1 A6000 \\
\midrule[0.3pt]
\multirow{2}{*}{\textbf{70B}} & N/A & 16 A100 & 2 A100 & 1 A100 \\
& N/A & 26 A6000 & 4 A6000 & 2 A6000 \\
\bottomrule
\end{tabular}
\caption{Minimum GPU requirements for A100~(80G) and A6000~(48G) when training models with different sizes under four situations. N/A denotes DDP cannot be applied for such large models.}
\label{tab:GPU}
\end{table}

\section{Utilization and Evaluation}  \label{sec:design-utilization}
After training, we can develop suitable prompting strategies to use LLMs and assess their effectiveness. 
Users can reuse existing models, APIs or the models trained by LLMBox.
The framework of our utilization pipeline is depicted in Figure~\ref{fig:framework}.

\subsection{Utilization Methods}
We include quantization deployment strategies for using LLMs alongside two prompting methods: in-context learning (ICL) and chain-of-thought (CoT).

$\bullet$ \emph{Quantization.}
To enhance memory efficiency during inference, LLMBox incorporates two quantization techniques: bitsandbytes~\cite{dettmers-etal-2022-bit} and GPTQ~\cite{frantar-etal-2023-optq}. Both methods facilitate 8-bit and 4-bit quantization and GPTQ additionally supports 3-bit quantization.

$\bullet$ \emph{In-context learning.}
We design a unified dataset class to organize diverse examples for few-shot learning. Furthermore, we implement three advanced ICL strategies, including KATE for example selection~\cite{liu-etal-2022-makes}, GlobalE for example order arrange~\cite{lu-etal-2022-fantastically}, and APE for instruction designing~\cite{zhou-etal-2023-large}.

$\bullet$ \emph{Chain-of-thought.}
Moreover, LLMBox incorporates several CoT prompting methods, such as program-aided (PAL) CoT~\cite{gao-etal-2023-pal} and least-to-most CoT~\cite{zhou-etal-2023-leasttomost}. We develop a flexible framework to facilitate self-consistency~\cite{wang-etal-2023-consistency} and repeated sampling~\cite{nijkamp-etal-2023-codegen}, which are beneficial for tasks involving mathematics and coding.

\subsection{Evaluation Methods}
In LLMBox, we implement the evaluation of LLM performance through three distinct mechanisms:

$\bullet$ \emph{Free-form generation:}
This is the basic evaluation method for generative LLMs and is applicable across all tasks. Models are required to generate responses to queries in an auto-regressive manner. We integrate common decoding strategies, including greedy search, temperature sampling, top-$p$ sampling, repetition penalties, \etc.

$\bullet$ \emph{Completion perplexity:}
This method is widely adopted for assessing multi-choice tasks in base LLMs. It involves comparing the perplexity (PPL) of each completion conditioned on the context and choose the one with the lowest average PPL. Additionally, we incorporate the use of normalized PPL as introduced in GPT-3~\cite{brown-etal-2020-language}.

$\bullet$ \emph{Option probability:}
Similar to the multi-choice formats in human examination, we feed a context with all the options to LLMs and require them to select the option letter (\eg A). This approach is commonly utilized in chat-based models.

Significantly, we introduce \emph{prefix caching} mechanism that caches the hidden states of common prefix texts to speed up the inference process. This strategy is organized at two levels: (1) we store the states of few-shot examples and compute them just once for all instances, \eg 5-shot examples in MMLU~\cite{hendrycks-etal-2021-measuring} and 8-shot examples in GSM8K~\cite{cobbe-etal-2021-training}; (2) we cache the states of identical contexts of different options when calculating completion perplexity. The effectiveness of this method is verified in Section~\ref{sec-efficiency}.

\subsection{Supported Models}
We integrate a variety of LLMs to keep pace with the swift advancements in this field. Given that LLMBox is based on the Transformers library~\cite{wolf-etal-2020-transformers}, it is compatible with a vast majority of publicly available models.
We list some included models as follows:

$\bullet$ \emph{General models:} LLaMA~\cite{touvron-etal-2023-llama} and Mistral~\cite{jiang-etal-2023-mistral};

$\bullet$ \emph{Chinese models:} Qwen~\cite{bai-etal-2023-qwen} and Baichuan~\cite{yang-etal-2023-baichuan};

$\bullet$ \emph{Multilingual models:} BLOOM~\cite{le-etal-2022-bloom};

$\bullet$ \emph{Chat models:} LLaMA-2 Chat~\cite{touvron-etal-2023-llama2} and Vicuna~\cite{chiang-etal-2023-vicuna};

$\bullet$ \emph{Code models:} CodeGen~\cite{nijkamp-etal-2023-codegen} and StarCoder~\cite{li-etal-2023-starcoder};

$\bullet$ \emph{Mathematical models:} Llemma~\cite{azerbayev-etal-2024-llemma} and DeepSeekMath~\cite{shao-etal-2024-deepseekmath}.

We also incorporate commercial APIs including OpenAI\footnote{\url{https://openai.com/}} and Anthropic Claude\footnote{\url{https://www.anthropic.com/}}.

\subsection{Supported Tasks}
Currently, LLMBox integrates 18 diverse tasks and corresponding 56 datasets with hundreds of subsets. 
The broad range of supported datasets within LLMBox enables to evaluate various models. For instance, users can employ English benchmarks, language modeling, and knowledge reasoning datasets for evaluating foundational pre-trained LLMs. In the case of chat-based models, users can utilize datasets focused on open-ended dialogue, human alignment, and instruction following.
We list some included tasks and datasets as follows:

$\bullet$ \emph{English benchmarks:} MMLU~\cite{hendrycks-etal-2021-measuring} and BBH~\cite{srivastava-etal-2023-beyond};

$\bullet$ \emph{Chinese benchmarks:} CMMLU~\cite{li-etal-2023-cmmlu} and C-Eval~\cite{huang-etal-2023-ceval};

$\bullet$ \emph{Multilingual benchmarks:} TyDi QA~\cite{clark-etal-2020-tydi} and MGSM~\cite{shi-etal-2023-language};

$\bullet$ \emph{Language modeling:} LAMBADA~\cite{paperno-etal-2016-lambada};

$\bullet$ \emph{Open-ended dialogue:} MT-Bench~\cite{zheng-etal-2023-judging} and AlpacaEval~\cite{li-etal-2023-alpaca};

$\bullet$ \emph{Machine translation:} general translation task in WMT\footnote{\url{https://www2.statmt.org/}} of every year and Flores-200~\cite{costa-etal-2022-no};
8

$\bullet$ \emph{Text summarization:} CNN/Daily Mail~\cite{see-etal-2017-get} and XSum~\cite{narayan-etal-2018-dont};

$\bullet$ \emph{Code synthesis:} HumanEval~\cite{chen-etal-2021-evaluating} and MBPP~\cite{austin-etal-2021-program};

$\bullet$ \emph{Closed-book question answering:} Natural Questions~\cite{kwiatkowski-etal-2019-natural} and TriviaQA~\cite{joshi-etal-2017-triviaqa};

$\bullet$ \emph{Reading comprehension:} SQuAD 2.0~\cite{rajpurkar-etal-2018-know} and RACE~\cite{lai-etal-2017-race};

$\bullet$ \emph{Knowledge reasoning:} HellaSwag~\cite{zellers-etal-2019-hellaswag} and ARC~\cite{clark-etal-2018-think};

$\bullet$ \emph{Symbolic reasoning:} Tables of Penguins~\cite{herzig-etal-2020-tapas} and Colored Objects~\cite{srivastava-etal-2023-beyond};

$\bullet$ \emph{Mathematical reasoning:} GSM8K~\cite{cobbe-etal-2021-training} and MATH~\cite{hendrycks-etal-2021-measuring};

$\bullet$ \emph{Human Alignment:} TruthfulQA~\cite{lin-etal-2022-truthfulqa} and CrowS Pairs~\cite{nangia-etal-2020-crows};

$\bullet$ \emph{Hallucination detection:} HaluEval~\cite{li-etal-2023-halueval};

$\bullet$ \emph{Instruction following:} IFEval~\cite{zhou-etal-2023-instruction};

$\bullet$ \emph{Environment Interaction:} ALFWorld~\cite{shridhar-etal-2021-alfworld} and WebShop~\cite{yao-etal-2022-webshop};

$\bullet$ \emph{Tool Manipulation:} Gorilla~\cite{patil-etal-2023-gorilla}.

\begin{table*}[!ht]
\centering
\small
\resizebox{1\textwidth}{!}{
\begin{tabular}{cc|ccccccccc}
\toprule
\multicolumn{2}{c}{\textbf{LLaMA-2}} & \textbf{MMLU} & \textbf{BBH} & \textbf{HumanEval} & \textbf{NQs} & \textbf{HellaSwag} & \textbf{ARC-C} & \textbf{WinoGrande} & \textbf{BoolQ} & \textbf{GSM8K} \\
\midrule
\multirow{2}{*}{\textbf{7B}} & \textbf{Paper} & 45.3 & 32.6 & 12.8 & 25.7 & 77.2 & 45.9 & 69.2 & 77.4 & 14.6 \\
& \textbf{LLMBox} & 46.5 & 33.2 & 13.6 & 25.5 & 75.6 & 49.6 & 69.6 & 78.5 & 14.6 \\
\midrule
\multirow{2}{*}{\textbf{70B}} & \textbf{Paper} & 68.9 & 51.2 & 29.9 & 39.5 & 85.3 & 57.4 & 80.2 & 85.0 & 56.8 \\
& \textbf{LLMBox} & 69.5 & 54.8 & 29.2 & 40.3 & 83.3 & 57.8 & 80.7 & 85.6 & 56.6 \\
\bottomrule
\end{tabular}}
\caption{The results of different tasks on LLaMA-2 (7B) and (70B).}
\label{tab:dataset}
\end{table*}
\section{Library Usage} \label{sec:usage}
In this section, we present the application of our library across four distinct research scenarios, illustrated through example code snippets.

\begin{figure}[!t]
    \centering
    \includegraphics[width=0.99\linewidth]{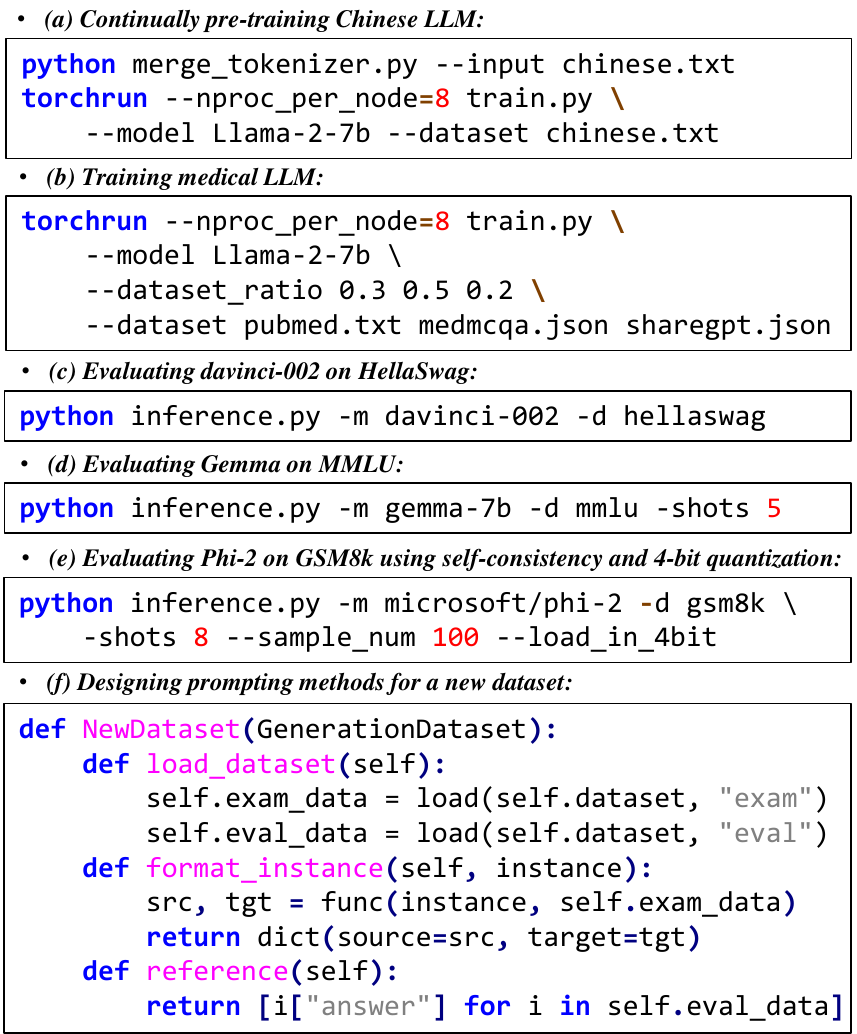}
    \caption{Usage examples of our LLMBox library on six representative tasks.}
    \label{fig:usage}
    \vspace{-0.2cm}
\end{figure}

\paragraph{Continually Pre-Training Language-Specific Models.}
As introduced in Section~\ref{sec:design-training}, we facilitate the continual pre-training of existing English-based LLMs for quick acquisition of new languages. Figure~\ref{fig:usage} (a) illustrates the process of tuning a Chinese LLM from LLaMA-2. Users are required only to prepare Chinese plain texts, such as those from Wikipedia, into a file named \texttt{chinese.txt}. Subsequently, LLMBox integrates new Chinese tokens into the vocabulary and trains the model.

\paragraph{Adapting LLMs to Specialized Domains.}
LLMBox facilitates the adaptation of LLMs to various specialized domains via instruction tuning, covering domains such as medicine, law, and finance. We present a script in Figure~\ref{fig:usage} (b) to train a medical LLM. We implement a convenient dataset mixture approach to sample instances from raw medical texts, medical instruction data, and general conversation data. This enables users to adjust the proportion to make a balance between medical knowledge, medical tasks, and conversational skills, thereby crafting an effective medical assistant.

\paragraph{Comprehensively Evaluating LLMs.}
We cover a broad range of tasks and various models within LLMBox to implement comprehensive evaluation. As illustrated in Figure~\ref{fig:usage} (c), (d), and (e), we present three exemplary command lines. Users are only required to designate the model and dataset names via the \texttt{-m} and \texttt{-d} options to achieve an efficient and accurate assessment of model performance. Furthermore, LLMBox supports multiple utilization methods, such as in-context learning (\texttt{-shots}), self-consistency (\texttt{-{}-sample\_num}), and quantitation (\texttt{-{}-load\_in\_4bit}).

\paragraph{Designing Novel Prompting Methods.}
Since the implementation of each dataset in LLMBox is unified, it offers the flexibility to add new datasets or design various prompting methods without affecting other modules. Figure~\ref{fig:usage} (f) overviews the design of our \texttt{Dataset} class. When adding a new dataset, users are only required to implement three functions: \texttt{load\_dataset} to load evaluation and example datasets; \texttt{format\_instance} to format each instance with instruction or few-shot examples; and \texttt{reference} to define the ground truth. In the core function \texttt{format\_instance}, users can develop innovative prompting methods tailored for each evaluation instance using example datasets.

\section{Experiment} \label{sec:exp}
In the section, we conduct extensive experiments to verify the effectiveness and efficiency.

\subsection{Effectiveness Evaluation}
The essential attribute of an open-source library is its ability to reproduce results effectively. To confirm this, we choose several representative training and utilization scenarios for testing the outcomes derived from LLMBox.

\begin{table}[!t]
\centering
\small
\begin{tabular}{c|cc}
\toprule
\textbf{Proportion} & \multirow{2}{*}{\textbf{MMLU}} & \multirow{2}{*}{\textbf{Alpaca-Eval}} \\
\textbf{FLAN / Alpaca} &  &  \\
\midrule
100 / 0 & 50.6  & 15.0 \\
50 / 50 & 50.5  & 44.4  \\
0 / 100 & 47.5  & 47.2  \\
\cmidrule{1-3}
LLaMA-2 (7B) & 46.5 & 23.0 \\
\bottomrule
\end{tabular}
\caption{The performance of base LLaMA-2 (7B) and instruction tuned results using different data mixture.}
\label{tab:sft_proportion}
\end{table}

\paragraph{Training results.}
We train LLaMA-2~\cite{touvron-etal-2023-llama2} with the mixture of instruction tuning data FLAN~\cite{chung-etal-2022-scaling} and Alpaca-52K~\cite{taori-etal-2023-alpaca} and evaluate its performance. We adjust the proportions of these datasets and assess the impact on performance using the MMLU benchmark~\cite{hendrycks-etal-2021-measuring} and the chat-oriented AlpacaEval~\cite{dubois-etal-2023-alpacafarm}. The experiments are conducted with a batch size of 128 and a constant learning rate of $1 \times 10^{-5}$. The model undergoes training for a total of 1200 steps, and we report the peak performance observed on the evaluation datasets.
The results in Table~\ref{tab:sft_proportion} indicate that FLAN improves the model's performance on NLP tasks, whereas Alpaca-52K significantly enhances its performance in daily chat. Moreover, when mixing both instruction datasets yields a balanced improvement across both tasks, aligning with findings from prior research~\cite{wang-etal-2023-how}.

\begin{table}[!t]
\centering
\small
\begin{tabular}{c|ccc}
\toprule
\textbf{Models} & \textbf{HellaSwag} & \textbf{MMLU} & \textbf{GSM8K} \\
\midrule
\textbf{GPT-NeoX} (20B)     & 71.4 & 26.4 & 7.1 \\
\textbf{OPT} (66B)          & 73.5 & 27.3 & 2.2 \\
\textbf{BLOOM} (7.1B)       & 61.1 & 26.0 & 4.2 \\
\textbf{LLaMA-2} (70B)      & 83.4 & 69.5 & 56.7 \\
\textbf{Pythia} (12B)       & 67.2 & 25.1 & 4.6 \\
\textbf{MPT} (30B)          & 79.8 & 45.4 & 21.5 \\
\textbf{Phi-2} (2.7B)       & 73.1 & 57.7 & 55.5 \\
\textbf{Mistral} (7B)       & 80.2 & 63.8 & 43.6 \\
\textbf{Falcon} (40B)       & 82.5 & 56.4 & 27.1 \\
\textbf{Gemma} (7B)         & 79.2 & 65.3 & 52.3 \\
\bottomrule
\end{tabular}
\caption{The results of different English LLMs using our developed LLMBox.}
\label{tab:english}
\end{table}

\paragraph{Utilization results.}
Firstly, we examine the performance of LLaMA-2~\cite{touvron-etal-2023-llama2} across various supported tasks. We totally evaluate nine tasks, including MMLU (5-shot, accuracy)~\cite{hendrycks-etal-2021-measuring}, BBH (3-shot, accuracy)~\cite{srivastava-etal-2023-beyond}, HumanEval (0-shot, pass\@1)~\cite{chen-etal-2021-evaluating}, Natural Questions~(NQs, 5-shot, EM)~\cite{kwiatkowski-etal-2019-natural}, HellaSwag (0-shot, accuracy)~\cite{zellers-etal-2019-hellaswag}, ARC-Chanllge~(ARC-C, 0-shot, accuracy)~\cite{clark-etal-2018-think}, WinoGrande (0-shot, accuracy)~\cite{Sakaguchi-etal-2021-winogrande}, BoolQ (0-shot, accuracy)~\cite{clark-etal-2019-boolq}, and GSM8K (8-shot, accuracy)~\cite{cobbe-etal-2021-training}. The results in Table~\ref{tab:dataset} demonstrates that our LLMBox library faithfully reproduces the results reported in their original papers. Furthermore, we verify the performance of LLMBox across a variety of models. We utilize HellaSwag, MMLU, and GSM8K to evaluate the performance of ten English LLMs, including GPT-NeoX~\cite{black-etal-2022-gpt}, OPT~\cite{zhang-etal-2022-opt}, BLOOM~\cite{le-etal-2022-bloom}, LLaMA-2~\cite{touvron-etal-2023-llama2}, Pythia~\cite{biderman-etal-2023-pythia}, MPT~\cite{mosaicml-etal-2023-introducing}, Phi-2~\cite{javaheripi-etal-2023-phi}, Mistral~\cite{jiang-etal-2023-mistral}, Falcon~\cite{almazrouei-etal-2023-falcon}, Gemma~\cite{googleteam-etal-2024-gemma}. We employ HellaSwag, C-Eval~\cite{huang-etal-2023-ceval}, and GSM8K to evaluate the performance of eight Chinese LLMs, including ChatGLM3~\cite{zeng-etal-2022-glm}, Chinese-LLaMA-2~\cite{cui-etal-2023-efficient}, InternLM-2~\cite{internlm-etal-2023-internlm}, Baichuan-2~\cite{baichuan-etal-2023-baichuan2}, Qwen-1.5~\cite{bai-etal-2023-qwen}, Aquila-2~\cite{BAAI-etal-2024-aquila2}, Deepseek~\cite{deepseekai-etal-2024-deepseek}, Yi~\cite{young-etal-2024-yi}. The results of these evaluations are detailed in Tables~\ref{tab:english} and~\ref{tab:chinese}. We can find that our LLMBox is also compatible with various English and Chinese LLMs.

\begin{table}[!t]
\centering
\small
\begin{tabular}{c|ccc}
\toprule
\textbf{Models} & \textbf{HellaSwag} & \textbf{C-Eval} & \textbf{GSM8K} \\
\midrule
\textbf{ChatGLM-3} (6B)	        & 63.6	& 53.0	& 48.5   \\
\textbf{C-LLaMA-2} (13B)	    & 76.4	& 41.8	& 18.6   \\
\textbf{InternLM-2} (20B)	    & 82.5	& 69.5	& 74.4	   \\	
\textbf{Baichuan-2} (13B)	    & 74.7	& 59.2	& 42.8   \\		
\textbf{Qwen-1.5} (72B)	        & 83.8	& 83.5	& 78.2   \\
\textbf{Aquila-2} (34B)	        & 78.8	& 98.6	& 2.0   \\
\textbf{Deepseek} (67B)	        & 83.4	& 65.9	& 64.1   \\
\textbf{Yi} (34B)		        & 83.2	& 81.4	& 5.4   \\
\bottomrule
\end{tabular}
\caption{The experimental results of different Chinese LLMs and APIs using our developed LLMBox. C-LLaMA-2 is short for Chinese-LLaMA-2.}
\label{tab:chinese}
\end{table}

\subsection{Efficiency Evaluation} \label{sec-efficiency}
The implementation efficiency is also a key factor to deploy LLMs. In addition to accurately reproducing results, we have optimized LLMBox for training and utilization efficiency.
From the results in Table~\ref{tab:efficiency}, it is evident that our prefix caching approach substantially decreases the inference time compared to the traditional Transformers implementation. As the number of examples increases (from 5-shot setting in MMLU to 8-shot setting in GSM8K), the efficiency gains from our method become increasingly pronounced. Remarkably, with the application of our prefix caching technique to the MMLU benchmark, LLMBox requires merely six minutes to process over ten thousand instances, achieving a 60\% reduction in processing time compared to the vLLM toolkit. In the future, we aim to incorporate this prefix caching strategy into vLLM to further enhance the inference efficiency.

\begin{table}[!t]
\centering
\small
\begin{tabular}{c|ccc}
\toprule
\textbf{Strategies} & \textbf{HellaSwag} & \textbf{MMLU} & \textbf{GSM8K} \\
\midrule
\textbf{Transformers}    & 5.5 & 18.5 & 130.5 \\
\textbf{Transformers+PC} & 6.1 & 6.0 & 23.3 \\
\textbf{vLLM}   & 6.6 & 14.9 & 3.6 \\
\bottomrule
\end{tabular}
\caption{The execution time of different implementation methods on LLaMA-2 (7B) using one A800 (80G) GPU. PC is short for the proposed novel prefix caching mechanism in our developed LLMBox.}
\label{tab:efficiency}
\end{table}

\section{Conclusion}\label{sec:conclusion}
This paper presented \textbf{LLMBox}, a comprehensive library for conducting research on training, utilizing, and evaluating large language models. For training, we designed a unified data interface to support the implementation of various training strategies. For utilization and evaluation, we implemented typical approaches to use LLMs (including quantization, ICL, and CoT prompting), covered 18 tasks and 56 datasets, and included a number of popular open-sourced LLMs and closed-source APIs. We further conducted extensive experiments to verify the effectiveness and efficiency of LLMBox.  Our library provides a unified framework to compare, reproduce, and develop LLMs and supporting methods for academic purposes, which would be of important value to promote the research on LLMs.

\section*{Acknowledgement}
This work was partially supported by National Natural Science Foundation of China under Grant No. 62222215, Beijing Natural Science Foundation under Grant No.4222027 and L233008. Xin Zhao is the corresponding author.
  
\bibliography{ref}

\end{document}